\documentclass[11pt]{article}

\usepackage{acl}
\usepackage{times}
\usepackage{latexsym}
\usepackage[T1]{fontenc}
\usepackage[utf8]{inputenc}
\usepackage{microtype}
\usepackage{inconsolata}
\usepackage{graphicx}
\usepackage{amsmath}
\usepackage{amssymb}
\usepackage{booktabs}
\usepackage{multirow}
\usepackage{array}

\title{A Declarative--Procedural Perspective on Expert Routing in Bilingual Mixture-of-Experts Language Models}

\author{
Amrit Gopinath$^{1}$ \and Raghul$^{1}$ \and Durairaj Thenmozhi$^{2}$ \\
$^{1}$Sri Sivasubramaniya Nadar College of Engineering, Chennai, India \\
$^{2}$Shiv Nadar University Chennai, India \\
\texttt{amrit2410182@ssn.edu.in, raghul2510435@ssn.edu.in} \\
\texttt{thenmozhid@snuchennai.edu.in}
}

\begin{document}

\maketitle

\begin{abstract}
We investigate whether Mixture-of-Experts (MoE) language models develop linguistically structured expert routing during bilingual language acquisition. Inspired by the Declarative–Procedural framework, we analyze lexical, grammatical, and syntactic processing in a decoder-only English–German MoE Transformer trained under sequential language exposure. We construct a probe-based validation set and extract token-level routing distributions to quantify category-dependent specialisation using mutual information, routing entropy, and Jensen–Shannon distance. The curriculum-trained model exhibits a peak mutual information of 0.1148 at layer 5, indicating category-dependent differences in routing distributions across linguistic categories. Surprisingly, a no-curriculum baseline trained on mixed English–German data shows stronger aggregate specialisation, reaching a peak mutual information of 0.2599 at the same layer. These results suggest that interpretable linguistic organization emerges within MoE routing patterns even without sequential language exposure. A replication at a second training seed shows that the no-curriculum condition's specialisation concentrates on a single language whose identity is seed-dependent, whereas the curriculum consistently yields a stable, language-balanced routing profile; rather than uniformly increasing specialisation, staged bilingual exposure reduces single-language dominance. The official Github repository: \href{https://github.com/Amrit828/DP-Theory-MOE-Interpretability-Research}{Github link}
\end{abstract}

\section{Introduction}

Mixture-of-Experts (MoE) language models route each token across a subset of expert networks instead of sending every token through the same feed-forward network \cite{shazeer2017outrageously,fedus2022switch}. This makes MoE models useful not only for scaling, but also for studying how neural language models choose experts internally. If different kinds of linguistic tokens are routed to different experts, the routing mechanism may exhibit an organized form of expert distribution inside the model.

In this work, we examine whether bilingual MoE language models develop routing patterns that correspond to meaningful linguistic categories. We focus on three categories motivated by the Declarative--Procedural view of language: lexical knowledge, grammatical processing, and syntactic structure \cite{ullman2001neurocognition,ullman2001neuralbasis,ullman2020dpmodel}. In this paper, a probe is a token chosen from a sentence to be examined. Lexical probes are tokens that test word knowledge, such as irregular forms. Grammatical probes are tokens that test rule-based morphology and agreement. Syntactic probes are tokens that participate in clause structure and sentence organization. Instead of examining only the model's final predictions, we directly analyse the expert-routing distributions associated with these probe tokens.

We train a sparse bilingual transformer using a staged English--German curriculum. The training curriculum starts with English and gradually transitions to German, approximating structured second-language exposure and curriculum learning in bilingual settings \cite{bengio2009curriculum,platanios2019competence,zhang2019curriculum}. We compare this structured curriculum to an unstructured no-curriculum setting in which English and German are introduced without gradual cumulative exposure. This lets us examine whether MoE routing becomes organized by language and whether staged bilingual exposure affects how that organization is distributed across languages.

To measure the development of an organized routing schema, we use a held-out validation set with lexical, grammatical, and syntactic probes in both English and German. We examine the router's probability distribution over experts at each MoE layer for each probe token. We then measure the relationship between expert routing and linguistic category using mutual information, entropy, Jensen--Shannon divergence, and sentence-level permutation testing \cite{shannon1948mathematical,lin1991divergence,cover2006elements}. This complements recent work that studies behavioral declarative and procedural knowledge in datasets and large language models \cite{li2024metacognitive} by moving the analysis to the level of internal routing.

Our results show that MoE routing distributions contain measurable information about linguistic category membership: the category of a probe token can be partially inferred from its expert probability distribution because routing and category exhibit non-trivial mutual information. Lexical, grammatical, and syntactic probes are routed differently. The clearest specialisation appears in intermediate MoE layers, whereas later layers route more diffusely and show reduced mutual information. We also find that curriculum affects how specialisation is balanced between languages. Without a curriculum, routing specialisation concentrates heavily on one language, and this concentration is seed-dependent; the staged L1--L2 curriculum instead yields a language-balanced routing organization that is stable across seeds.

In general, this study shows that bilingual MoE routers develop a language-sensitive quantifiable expert-routing structure that reflects linguistically significant differences. These findings suggest that expert routing provides a useful mechanistic window into how bilingual language models allocate computation across different types of language processing.

The central motivation of this paper is to move beyond asking whether a bilingual MoE model performs well on a diagnostic task and instead ask how its internal routing mechanism organizes linguistic computation. Our explicit contribution is a routing-level analysis framework for testing whether expert allocation is sensitive to lexical, grammatical, and syntactic probe categories under staged bilingual exposure. By comparing a forward English--German curriculum with a no-curriculum baseline, we isolate how training order changes the distribution of routing specialisation across languages. This makes the study a mechanistic investigation of expert routing in bilingual MoE models, rather than a general benchmark of language-model accuracy.

\section{Related Work}

This section places the study in the context of four connected areas: the Declarative--Procedural view of language, the linguistic analysis of transformer models, expert routing in MoE architectures, and curriculum learning for bilingual exposure. Together, these areas lead to the main question of the paper: does expert routing in a bilingual MoE model become sensitive to lexical, grammatical, and syntactic differences? Also, does staged language exposure change that routing structure?

\paragraph{Declarative--Procedural theory.}
The Declarative--Procedural framework divides language knowledge into two main parts \cite{ullman2001neurocognition,ullman2001neuralbasis,ullman2020dpmodel}. Declarative knowledge relates to stored word knowledge, such as vocabulary and irregular forms. Procedural knowledge relates to rule-based grammar and compositional structure. In this paper, we use this framework only as a linguistic lens. We do not claim that MoE routers are the same as human memory systems.

\paragraph{Declarative and procedural knowledge in LLMs.}
Recent studies have looked at declarative and procedural knowledge in language models mainly through model outputs and task performance \cite{li2024metacognitive}. Our work explores this question within the model. Instead of just asking if the model provides the right answer, we investigate whether a sparse bilingual model allocates computation differently for varying types of probe tokens.

\paragraph{Transformer linguistic structure.}
Previous studies indicate that transformer layers do not all encode the same type of linguistic information \cite{vaswani2017attention,tenney2019bert,clark2019bert,elhage2021framework}. Intermediate layers often have clearer syntactic and semantic structures than very early or very late layers. This motivates our layer-wise analysis of MoE routing. We specifically question whether middle routed layers show stronger category-sensitive expert allocation.

\paragraph{Mixture-of-Experts routing.}
MoE models use routers to direct tokens to a subset of experts \cite{shazeer2017outrageously,fedus2022switch}. Earlier work has shown that experts can specialize based on token statistics, domains, routing design, and balancing constraints \cite{dai2024deepseekmoe,falke2026moeroutingtestbed,sun2026expertthreshold}. Our research stands out by focusing on bilingual linguistic probe categories and exploring how curriculum structure impacts the distribution of routing specialization across English and German.

\paragraph{Curriculum learning and bilingual exposure.}
Curriculum learning investigates how the order of training examples influences learning \cite{bengio2009curriculum,platanios2019competence,zhang2019curriculum}. In bilingual training, staged exposure is useful because it allows us to compare two scenarios: one where the model sees English first and German later, and another where both languages are mixed from the start. This comparison helps us assess whether the order of language exposure affects how routing specialization develops across languages.

\section{Routing Analysis Framework}
\label{sec:framework}

The model contains MoE blocks at the routed layers $L=\{1,3,5,7\}$, each with $N=8$ experts. Let $E$ denote the expert-routing random variable, a categorical variable whose values are the eight experts, and let $C$ denote the linguistic category random variable, taking the three (not binary) values $C = \{\textit{lexical}, \textit{grammatical}, \textit{syntactic}\}$. For a token $t$ at routed layer $l$, the router produces a probability distribution over the eight experts, denoted $r^{(l)}(t)$; throughout the paper this is the full post-softmax router distribution, taken \emph{before} top-$k$ masking, not the sparse renormalised vector used in the forward computation.

\paragraph{Probing mechanism.}
Probe tokens are annotated in each sentence using a parser (details in Methodology and Appendix~\ref{app:probe}). The routing vector for the probe token is extracted from the router during inference and used to compute the routing statistics below, independently for each routed layer.

We test $H_0: P(E|C) = P(E)$ (routing is independent of category) against $H_1: P(E|C) \neq P(E)$, equivalently $H_0: MI(E;C)=0$ versus $H_1: MI(E;C)>0$. This concerns the aggregate routing distribution, not individual tokens: the cumulative expert allocation of several thousand category probes can still match the global allocation, so $H_0$ is an empirically realisable regime rather than a strawman -- a random-routing control produces it almost exactly (MI $\approx 0.002$; Appendix~\ref{app:ablations}), and the load-balancing objective used during training explicitly pushes the router toward category-independent expert usage.

\subsubsection{Statistical Testing Framework}

The primary statistical tool used is mutual information between expert-routing distributions and linguistic category. Higher mutual information indicates a stronger association between the router’s expert probability mass and linguistic category.

The marginal routing probability assigned to expert $e$ is computed as

\begin{equation}
p(e)
=
\frac{
\sum_{\tau \in T}
r^{(l)}(\tau)[e]
}
{
\sum_{e'}
\sum_{\tau \in T}
r^{(l)}(\tau)[e']
}
\label{eq:pe}
\end{equation}

and the joint probability between expert $e$ and category $c$ is computed as

\begin{equation}
p(e,c)
=
\frac{
\sum_{\tau \in T_c}
r^{(l)}(\tau)[e]
}
{
\sum_{c'}
\sum_{e'}
\sum_{\tau \in T_{c'}}
r^{(l)}(\tau)[e']
}
\label{eq:pec}
\end{equation}

where $T$ denotes the complete set of probe tokens and $T_c$ denotes the subset belonging to category $c$.

Mutual information is then computed as

\begin{equation}
MI(E;C)
=
\sum_e \sum_c
p(e,c)
\log
\left(
\frac{p(e,c)}
     {p(e)p(c)}
\right)
\label{eq:mi}
\end{equation}

where $p(e)$ is the marginal routing probability assigned to expert $e$ and $p(c)$ is the marginal probability of the routed token belonging to category $c$. Robustness is assessed with a sentence-level permutation test at $p < 0.01$.

\subsubsection{Entropy Analysis}

Entropy is the second statistical tool, measuring how distributed routing is across experts: higher entropy indicates a more diffuse expert-usage distribution, lower entropy indicates concentration on a smaller subset of experts. Shannon entropy is given by
\begin{equation}
H(E) = - \sum_e p(e) \ln p(e)
\end{equation}
where \(p(e)\) is the marginal distribution of expert \(e\) being chosen; when computed for a specific category, \(p(e)\) is estimated only from that category's probe tokens, so category-wise entropy measures how concentrated or diffuse expert usage is per category.

Jensen--Shannon Divergence (JSD) measures how routing differs between linguistic categories, given by

\begin{equation}
\begin{aligned}
JSD\!\left(P(E|C_i)\;||\;P(E|C_j)\right) \\
=
\frac{1}{2}
D_{KL}\!\left(P(E|C_i)\;||\;M\right) \\
+
\frac{1}{2}
D_{KL}\!\left(P(E|C_j)\;||\;M\right)
\label{eq:jsd}
\end{aligned}
\end{equation}

where
\[
M = \frac{1}{2}\left(P(E|C_i) + P(E|C_j)\right).
\]

We report its square root, the \emph{Jensen--Shannon distance} $d_{JS} = \sqrt{JSD}$, throughout the paper: unlike the divergence itself, this is a true metric \cite{endres2003metric}, making the three pairwise category separations directly comparable.

\section{Methodology}

\subsection{Dataset Construction and Probe Annotation}

The dataset is constructed from the FineWeb-Edu corpus \cite{penedo2024fineweb} for the English part and the German portion of mC4 \cite{xue-etal-2021-mt5} for the German part. The pipeline uses the Stanza module to categorise words in each sentence as lexical, grammatical, and syntactic. The taxonomy for this classification is given in Appendix~\ref{app:probe}.

One probe token is annotated for each sentence-category pair. A sentence may contain probe tokens belonging to multiple linguistic categories and can therefore appear in more than one category-specific dataset. In such cases, the sentence is duplicated across the relevant categories, with each category using its corresponding probe token.

The following table provides the size of the English and German datasets:

\begin{table}[ht]
\centering
\label{tab:dataset_size}
\begin{tabular}{lccc}
\hline
Language & Lexical & Grammatical & Syntactic \\
\hline
English & 460,250 & 650,000 & 650,000 \\
German  & 138,127 & 175,000 & 175,000 \\
\hline
\end{tabular}
\caption{Dataset size by language and linguistic category.}
\end{table}

\textbf{Note:} Lexical probe extraction was terminated before reaching the target dataset size because of computational constraints, resulting in smaller lexical subsets for both languages. This does not affect the reported analyses, which are conducted on a fixed held-out validation set.

\subsection{Sequential L1--L2 Curriculum}

The curriculum is designed to approximate staged second-language acquisition. Here, the first language is English (EN) and the second language is German (DE). The first 1--8 epochs consist of an EN-only curriculum. From epochs 9--16, the curriculum slowly transitions from EN-only to bilingual (EN + DE). The choice of language was based on the differences in grammatical structure and the linguistic category taxonomy (more details in Appendix~\ref{app:probe}).

\begin{equation}
\lambda_s = 0.20 + 0.05(s - 9), \qquad 9 \leq s \leq 14
\end{equation}

where $\lambda_s$ denotes the proportion of German samples presented during epoch $s$.

Within each epoch, the amount of lexical, grammatical, and syntactic probe tokens is equal and follows a repetition + new exposure system. This system can be given by

\begin{equation}
\begin{aligned}
Cur_s
&= 0.6 \cdot Cur_{s-1} \\
&\quad + 0.4 \cdot (New \;\text{or}\; Cur_{s'} \text{ where } s' < s-1)
\end{aligned}
\label{eq:curriculum}
\end{equation}

for $2 \leq s \leq 16$.

\begin{equation}
Cur_1 = 1 \cdot New
\end{equation}

where $Cur_s$ is the curriculum of epoch $s$ and $New$ is the set of samples that are not part of any previous curriculum.

\subsection{Diagnostic Held-out Validation}

The validation set consists of 12,000 samples (6,000 per language and 2,000 per category per language). The validation set is a minimal-pairs dataset consisting of a pair of sentences, $S^{+}$ and $S^{-}$, where $S^{+}$ is a linguistically valid and acceptable sentence and $S^{-}$ is a linguistically invalid sentence.

The $S^{+}$ set was generated synthetically using OpenAI's GPT-5.1 model in order to obtain controlled, diverse, and task-specific examples. The generation was guided by rigorous prompting (prompts provided in Appendix~\ref{app:openai_generation}), with valid probe tokens injected into the prompts. Automatic post-generation validation is applied to ensure that generated sentences follow the correct category subtype, are not duplicates, and contain the required metadata. The $S^{-}$ sentence is then deterministically obtained by modifying the probe token to invalidate the sentence in a linguistically meaningful way (rules provided in Appendix~\ref{app:openai_generation}).

\subsection{Experimental Setup}

A custom 8-layer MoE language model is used, with a feed-forward dimension of 2048, an embedding size of 512, and 8 attention heads per layer. The model uses the multilingual mBERT tokenizer with a vocabulary size of 119,547. The total parameter count of the model is 86.4 million.

A relatively small model size is chosen to facilitate controlled experimentation and clearer analysis of expert routing behaviour. The MoE layers are placed in alternate transformer layers (1, 3, 5, and 7). The primary training setup uses the proposed sequential L1--L2 curriculum, with an additional No-Curriculum setting used as an ablation baseline.

The no-curriculum baseline uses a fixed 80:20 English-German mixture at every epoch. This keeps the overall bilingual composition broadly similar to the forward curriculum, although the aggregate exposure is not exactly identical: the forward curriculum yields 81.56\% English and 18.44\% German across the full schedule, whereas the no-curriculum condition remains fixed at 80\% English and 20\% German.

Both conditions receive the same total number of training samples. The difference between them is the order in which the languages are introduced. In the forward curriculum, the model is trained on English first and German is introduced gradually. In the no-curriculum condition, English and German are mixed from the start using the same overall English-German proportions.

The primary runs use top-$k$ routing with $k=3$ and a router load-balancing coefficient of 0.01. Additional hyperparameters are provided in Appendix~\ref{app:hparams}.

\section{Results and Analysis}

All the results, mathematical analyses, and metrics reported below are obtained through inference on the forward-curriculum model and the no-curriculum ablation model over the held-out validation set.
Unless otherwise stated, all MI values reported in Sections 5.1--5.5 are computed on the pooled bilingual validation set combining English and German probes. Language-specific analyses are introduced separately in Section 5.6.

\subsection{Behavioural Performance}

Table~\ref{tab:performance} reports validation performance for the forward-curriculum and no-curriculum conditions. The two setups achieve broadly comparable behavioural outcomes on the held-out diagnostic validation set. The Accuracy is computed as token-level next-token accuracy on the grammatical sentence strings only, averaged over all non-padding tokens; it is not a pairwise sentence-choice accuracy between $S^{+}$ and $S^{-}$. The forward-curriculum model attains lower perplexity (111.19) than the no-curriculum model (117.01), whereas the no-curriculum condition achieves marginally higher accuracy (29.65\% versus 28.99\%).

Because the two training conditions achieve similar behavioural performance, the routing patterns reported in the following sections are less likely to reflect simple differences in overall model competence.

The relatively low absolute scores reflect the challenging evaluation setup. The purpose of the diagnostic set is to expose routing behaviour under controlled linguistic contrasts rather than to serve as a benchmark of language-model capability. These scores are not the main outcome of the paper. They just indicate that both models can handle the diagnostic examples. The primary analysis focuses on the router. We examine whether different token types are sent to different experts during validation.

\begin{table}[ht]
\centering
\small
\begin{tabular}{lcc}
\toprule
Condition & Perplexity & Accuracy (\%) \\
\midrule
Forward Curriculum & 111.19 & 28.99 \\
No Curriculum & 117.01 & 29.65 \\
\bottomrule
\end{tabular}
\caption{Validation performance on the held-out diagnostic probe set.}
\label{tab:performance}
\end{table}

\subsection{Emergence of Category-Dependent Routing}

The primary analysis investigates whether routing specialisation occurs across different linguistic categories. Mutual Information (MI) is computed from the inference routing logs to quantify this phenomenon. The MI values vary across layers, indicating that the degree of specialisation differs throughout the network. Layer 5 shows the highest mutual information value (0.1148), suggesting the strongest category-dependent routing behaviour. In contrast, the final routed layer (layer 7) shows an MI value substantially lower than the other routed layers.

\begin{figure}[ht]
    \centering
    \includegraphics[width=\linewidth]{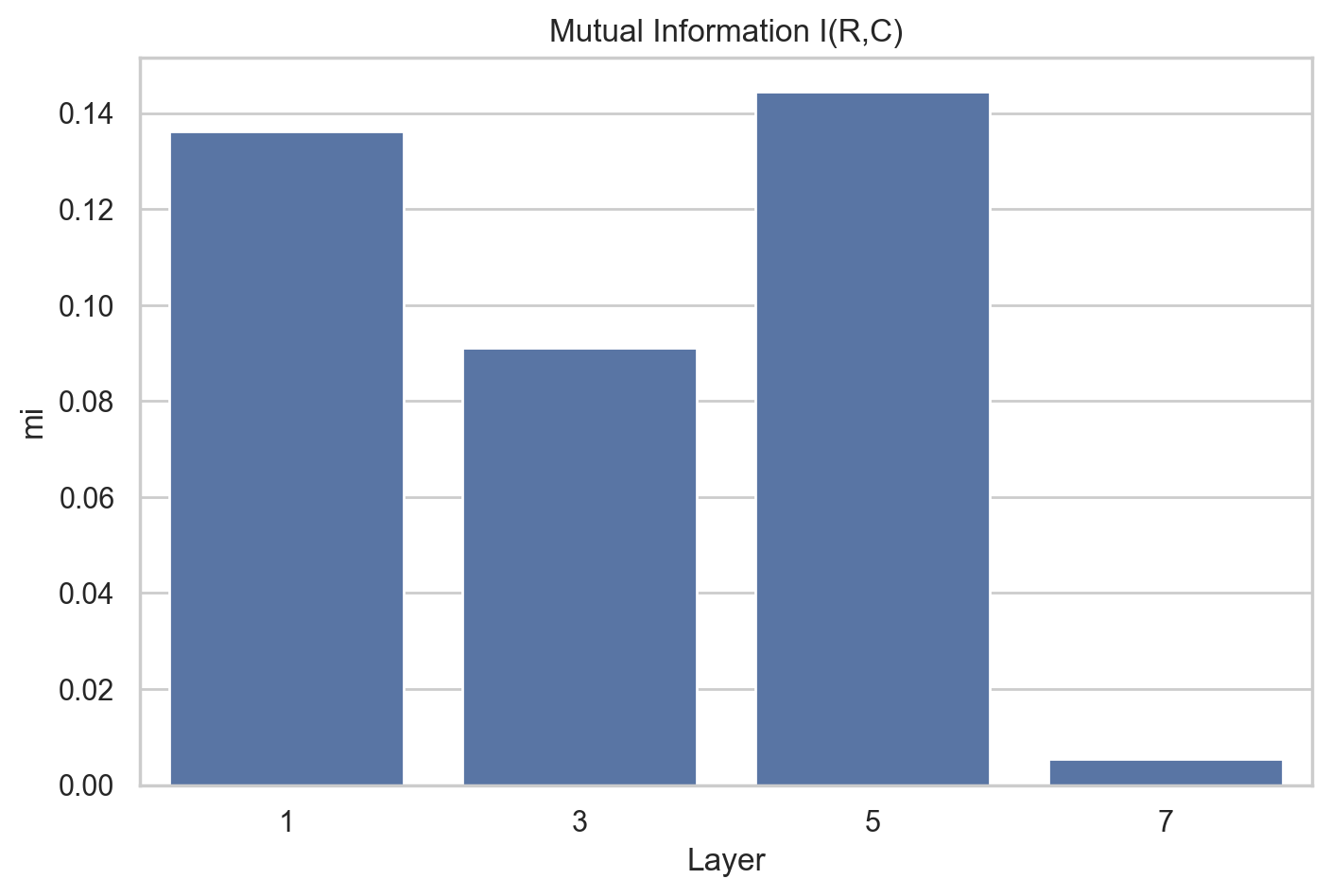}
    \caption{Layer-wise mutual information between expert-routing distributions and linguistic categories.}
    \label{fig:mi}
\end{figure}

To determine whether the obtained MI values (routing-category associations) are statistically significant, a permutation test was conducted using 1,000 sentence-level permutations. Across all routed layers, the observed MI values were significantly higher than the corresponding layer-wise null distributions, with statistical significance of $p < 0.001$ for all layers. This result makes a shuffled-label explanation unlikely and supports the presence of category-dependent routing behaviour.

For context, Appendix~\ref{app:ablations} shows a frozen (untrained) router reaches MI 0.0274 and random routing reaches $\approx$0.002 at layer 5: most of the trained model's 0.1148 requires learned routing, but the frozen baseline is not negligible (roughly a quarter of the effect) and should not be attributed entirely to router learning.

\begin{table}[ht]
\centering
\small
\begin{tabular}{ccccc}
\toprule
Layer & Observed MI & Null Mean & Null Std & $p$-value \\
\midrule
1 & 0.0520 & 0.0004 & 0.0002 & 0.000999 \\
3 & 0.0384 & 0.0003 & 0.0001 & 0.000999 \\
5 & 0.1148 & 0.0003 & 0.0001 & 0.000999 \\
7 & 0.0207 & 0.0001 & 0.0000 & 0.000999 \\
\bottomrule
\end{tabular}
\caption{Permutation test results for routing-category mutual information. All routed layers exhibit statistically significant routing-category associations. All reported p-values correspond to the minimum attainable empirical value under 1000 permutations, indicating that no permuted sample exceeded the observed MI.}
\label{tab:mi_perm}
\end{table}

\subsection{Expert Allocation Patterns}

The secondary analysis supporting the expert specialisation claim is performed through category-wise and layer-wise expert allocation patterns. This analysis also explains how experts have been utilized across different linguistic categories. Although the probability mass is distributed across different experts for each category, certain experts consistently receive higher routing probabilities for specific categories. This indicates the presence of preferential expert allocation rather than uniform routing.

\begin{figure}[ht]
    \centering
    \includegraphics[width=\linewidth]{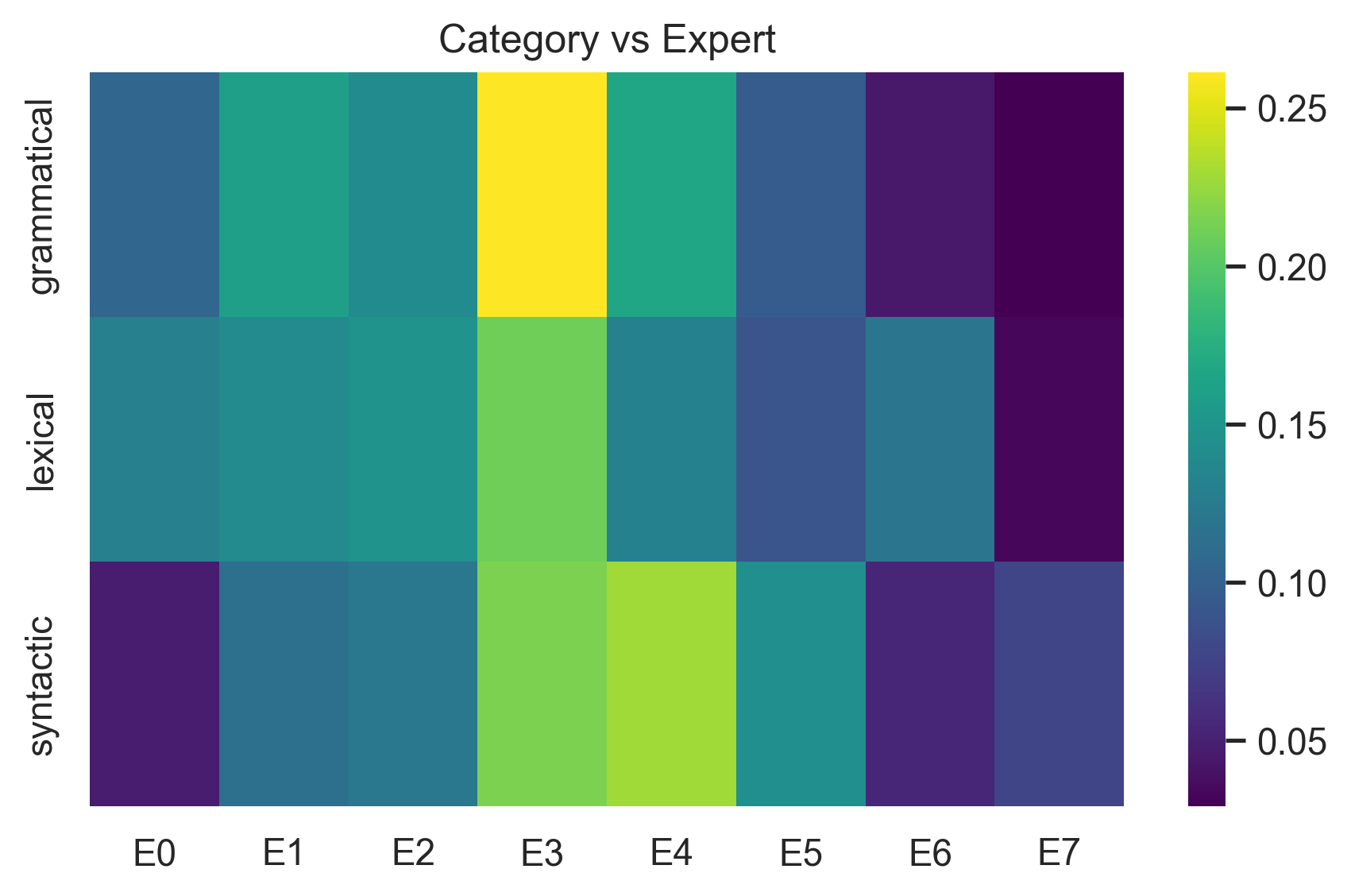}
    \caption{Category-conditioned expert allocation patterns.}
    \label{fig:heatmap}
\end{figure}

We can also observe that there is no collapse to a single expert for any particular linguistic category. Instead, each category exhibits a distribution of expert preferences, where some experts are selected more frequently than others. For example, grammatical probes show stronger preferences towards Experts E1 and E3, while syntactic probes exhibit higher utilisation of Experts E3--E5. This shows that category-dependent specialisation emerges through differences in routing preference rather than strict expert exclusivity.

\begin{figure}[ht]
    \centering
    \includegraphics[width=\linewidth]{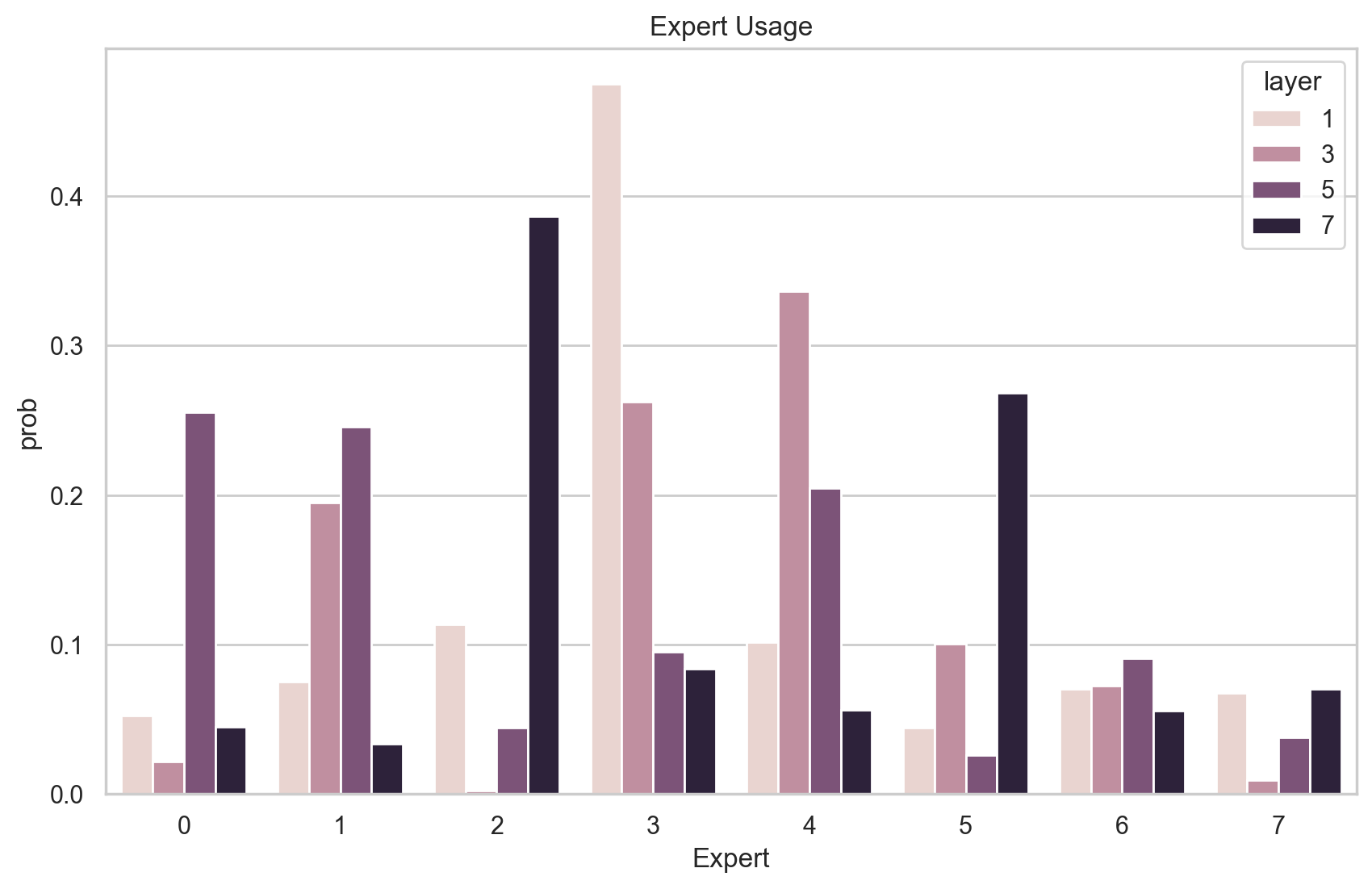}
    \caption{Layer-wise expert utilisation across routed layers.}
    \label{fig:expert_usage}
\end{figure}

Figure~\ref{fig:expert_usage} presents the layer-wise expert usage distribution. It is interesting to note that the different routed layers exhibit different expert usage distributions, with different subsets of experts dominating at different stages of the processing. This observation is consistent with the layer-wise MI analysis, suggesting that the routing distribution differs across different parts of the model.

While the heatmap and expert allocation reveal how the experts are allocated layer-wise as well as category-wise, they do not indicate how diverse or concentrated the routing distribution is. To examine this aspect, we analyse routing entropy across the routed layers.

\subsection{Routing Entropy Analysis}

The entropy values vary across different layers and among the different categories as well. Grammatical probes exhibit the lowest entropy consistently in all the layers (with the strongest effect in layer 3), which suggests that the expert utilisation distribution is more concentrated with a smaller subset of experts consistently utilized for this category.

In contrast, the syntactic probes exhibit the highest entropy values across different layers (or equal to lexical in layer 1). This indicates that expert usage is spread across more experts for this particular category across all layers. Together, these observations suggest that different linguistic categories rely on distinct routing strategies, with grammatical processing exhibiting more concentrated expert utilisation and syntactic processing exhibiting more distributed expert usage.

\begin{figure}[ht]
    \centering
    \includegraphics[width=\linewidth]{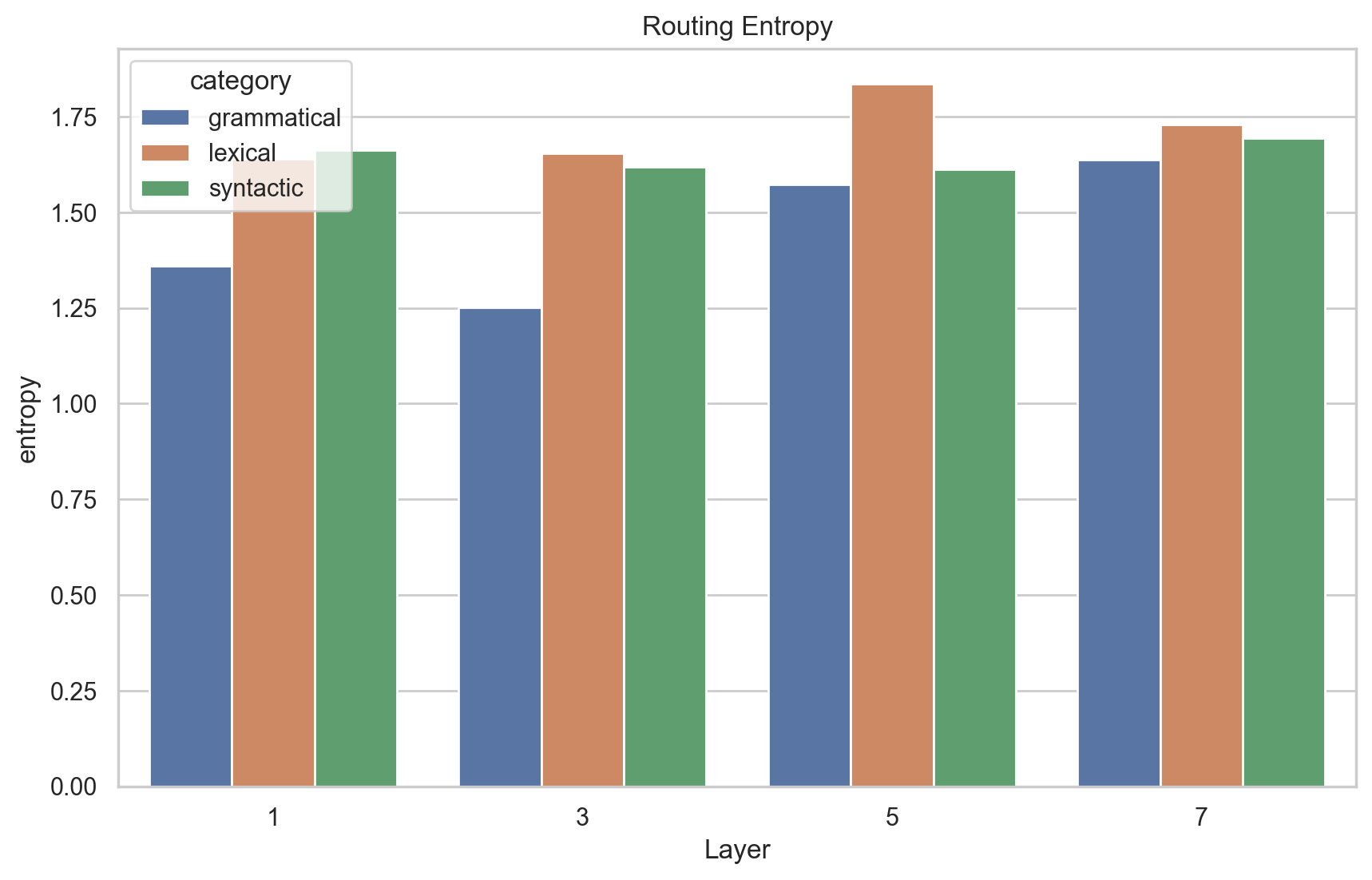}
    \caption{Routing entropy across routed layers and linguistic categories. Lower entropy indicates more concentrated expert utilisation.}
    \label{fig:entropy}
\end{figure}

\subsection{Category Separation Analysis}

While entropy explains how concentrated or distributed the expert utilisation is, it does not explain how the routing distribution differs between different linguistic categories. To examine this aspect, we compute the pairwise Jensen--Shannon distance (Section~\ref{sec:framework}) between category-wise routing distributions.

\begin{table}[h]
\centering
\small
\begin{tabular}{lcc}
\toprule
Left Category & Right Category & $d_{JS}$ \\
\midrule
Grammatical & Lexical & 0.0788 \\
Grammatical & Syntactic & 0.1147 \\
Lexical & Syntactic & 0.1218 \\
\bottomrule
\end{tabular}
\caption{Pairwise Jensen--Shannon distance between category-conditioned routing distributions.}
\label{tab:jsd}
\end{table}

Table~\ref{tab:jsd} shows the pairwise Jensen--Shannon distances between grammatical, lexical and syntactic routing distributions. All category pairs exhibit non-zero separation, suggesting that the router assigns different expert utilisation distributions to different linguistic categories. We can observe that the routing distributions are farthest apart between lexical and syntactic processing, whereas grammatical and lexical processing are closest. These observations are consistent with the mutual information and expert allocation analyses, providing additional evidence for category-dependent routing.

\subsection{Effect of Curriculum on Specialisation}

In the no-curriculum setting, the mutual information peaks at 0.2599 in layer 5 and reaches 0.1945 in layer 1. Similar to the forward curriculum setting, the strongest routing-category association is observed in the intermediate routed layers. These observations indicate that routing specialisation emerges under both training conditions, suggesting that curriculum learning is not strictly necessary for category-dependent routing to develop.

\begin{table}[h]
\centering
\small
\begin{tabular}{lc}
\toprule
Layer & No-Curr MI \\
\midrule
1 & 0.1945 \\
3 & 0.0674 \\
5 & 0.2599 \\
7 & 0.0739 \\
\bottomrule
\end{tabular}
\caption{Pooled bilingual routing-category MI under the no-curriculum condition.}
\end{table}

The following table shows language-wise MI values computed separately for English and German.

\begin{table}[h]
\centering
\small
\begin{tabular}{lcccc}
\toprule
Layer & Fwd EN & Fwd DE & No-Curr EN & No-Curr DE \\
\midrule
1 & 0.0508 & 0.1361 & 0.3810 & 0.1357 \\
3 & 0.0366 & 0.0911 & 0.1364 & 0.0804 \\
5 & 0.1025 & 0.1444 & 0.6457 & 0.1116 \\
7 & 0.0571 & 0.0053 & 0.1141 & 0.0649 \\
\bottomrule
\end{tabular}
\caption{Language-wise routing-category mutual information for the forward curriculum and no-curriculum settings.}
\label{tab:curriculum_lang}
\end{table}

To verify that the observed category-sensitive routing is not reducible to language identity alone, we additionally compute conditional mutual information \(I(R;C \mid L)\), where \(L\) denotes language identity. This analysis measures whether routing still contains information about lexical, grammatical, and syntactic category membership after conditioning on whether the probe is English or German. As reported in Appendix Table~\ref{tab:conditional_mi}, conditional mutual information remains non-zero across routed layers in both training conditions, indicating that the routing-category relationship is not explained solely by English-German separation.

Table~\ref{tab:curriculum_lang} shows the language-wise mutual information for each routed layer under both the forward curriculum and no-curriculum settings. The language-wise MI values are drastically different for English. In particular, at layer 5, the MI value increases from 0.1025 in the forward curriculum setting to 0.6457 in the no-curriculum setting, representing an increase of approximately 6.3 times. A similar trend can be observed across all routed layers.

In contrast, the difference between the category-dependent routing effect of the two setups is much less visible in German, where MI remains around 0.1 at layers 1 and 5 for both setups (0.1444 in the forward curriculum setting and 0.1116 in the no-curriculum setting at layer 5 for German language). A similar effect is observed in the other layers, where the differences remain comparatively small. A paired bootstrap over the identical 12,000 validation records (2,000 resamples) finds this German layer-5 difference statistically reliable ($\Delta\mathrm{MI} = +0.0328$, 95\% CI $[+0.0211, +0.0446]$, $p = 0.001$), while English MI is significantly higher without the curriculum at every layer.

\paragraph{Seed replication.}
Because the results above come from a single run per condition, we replicated both conditions at a second seed (full values in Appendix~\ref{app:ablations}, Table~\ref{tab:seed_replication}). The structural findings reproduce (middle-layer concentration, final-layer collapse, higher pooled MI without the curriculum at every layer), but the per-language composition does not: at the second seed the no-curriculum model concentrates on German rather than English (DE layer-5 MI 0.213 vs.\ EN 0.078), the reverse of the seed-42 pattern, so \emph{which} language dominates is seed-dependent and the per-language contrasts above should not be generalised across runs. What is robust across both seeds is the asymmetry itself and its remedy: the no-curriculum condition is dominated by a single language ($|$EN$-$DE$|$ MI at layer 5: 0.53 and 0.13), whereas the forward curriculum yields a balanced profile (0.04 and 0.08) and a more stable pooled MI (0.115 and 0.089, versus 0.260 and 0.113 without the curriculum).

Curriculum therefore does not uniformly increase specialisation or consistently favour either language: uncurriculated training acts as an initialization lottery handing specialisation to one language, and staged L1--L2 exposure converts this into a stable, language-balanced profile -- a reduction in single-language dominance rather than a language-specific improvement.

\section{Discussion}

Routing distributions provide significant insight into linguistic category membership, and specialisation appears in both training conditions, so curriculum learning is not necessary for category-sensitive routing to emerge; the open question is how curriculum changes its distribution across languages and layers. The strongest routing-category relationship occurs in intermediate layers, particularly layer 5, consistent with prior work suggesting intermediate transformer layers carry clearer linguistic structure, while the final routed layer shows lower mutual information and higher entropy, implying routing becomes more evenly spread across experts rather than category-separated.

The comparison between conditions is where pooled statistics can mislead: combining English and German into one score makes the no-curriculum model look stronger, but per-language analysis shows this pooled signal is a single-language phenomenon whose beneficiary is decided by the random seed, whereas the forward curriculum yields a language-balanced profile that is stable across seeds. Sequential L1--L2 exposure therefore does not simply boost or suppress specialisation; its seed-robust effect is to make the cross-lingual organization of specialisation predictable, i.e., to reduce single-language dominance rather than to redistribute specialisation toward a particular language.

Interpreting the behavioural accuracy values requires considering the study design: the model is intentionally small and the validation set uses controlled minimal-pair contrasts, so the aim is not high benchmark accuracy but comparable internal routing patterns across probe families; the behavioural results mainly indicate both models are sufficiently functional for meaningful routing analysis.

\section{Conclusion}

Bilingual MoE routing develops measurable category-dependent organization across lexical, grammatical, and syntactic probe families. The strongest routing specialisation is observed in intermediate routed layers and is robust on held-out validation data and across two training seeds. Curriculum learning does not uniformly increase specialisation; its seed-robust effect is to reduce single-language dominance rather than to favour a specific language. Without a curriculum, routing specialisation concentrates on a single language whose identity varies with the seed; staged L1--L2 exposure converts this initialization lottery into a stable, language-balanced bilingual specialisation profile.

\section*{Limitations}

This study uses a single EN--DE language pair and a single primary sparse architecture. While the observed routing patterns are consistent across multiple analyses, including mutual information, entropy, Jensen--Shannon distance, permutation testing, and conditional mutual information controls, the extent to which the findings generalize to other language families, larger-scale MoE architectures, or different curriculum schedules remains an open question. The main conditions are replicated at two training seeds; per-language MI magnitudes vary across seeds, and we accordingly claim only the seed-stable structure (middle-layer concentration, final-layer collapse, and the curriculum's language-balancing effect), not per-language magnitudes.

The diagnostic validation set is generated through a controlled LLM-assisted pipeline and subsequently validated using parser-based checks and deterministic probe construction rules. Although these procedures improve consistency and coverage, the evaluation set does not undergo manual human validation. Future work could incorporate expert human review and naturally occurring linguistic examples to further verify that the observed routing patterns generalize beyond synthetic diagnostic probes. The validation set is dominated by single-piece probes under the mBERT tokenizer, but full subword-aggregation robustness was not recomputed for the final archived routing logs used in this draft

The present study uses distinct source corpora for English and German. Although language-conditioned analyses remain significant, future work should examine matched-domain multilingual corpora.

A reverse curriculum is not included because the forward curriculum is explicitly designed around a fixed cumulative English-German exposure ratio. Simply reversing the schedule would alter not only the temporal order of language presentation but also the total exposure received by each language across training. Consequently, a naive DE→EN reversal would confound sequencing effects with differences in cumulative language exposure. A fair reverse-curriculum comparison would therefore require a separately constructed schedule that preserves overall language proportions while reversing the order of introduction.

Several potential confounds were explicitly examined. Routing-category associations remained stable across alternative subword aggregation strategies and remained non-zero after conditioning on language identity and coarse lexical frequency controls. Nevertheless, the present study does not fully disentangle all possible interactions between linguistic category, token identity, lexical frequency, morphological complexity, and part-of-speech information. As a result, the reported routing effects should be interpreted as category-sensitive routing behaviour rather than evidence of perfectly isolated category-specific mechanisms.

A stricter frequency-balanced lexical control was explored but was not included in the final analysis. The curriculum-generated lexical distributions were highly skewed, making it difficult to construct well-matched irregular and regular subsets while maintaining sufficient probe coverage and sample diversity.

We do not yet report extensive architectural sweeps, multiple language pairs, or broad-scale hyperparameter sensitivity analyses. In addition, expert intervention experiments (e.g., expert masking or routing interventions) are not included, limiting our ability to make strong causal claims about the functional role of individual experts.

Finally, the work focuses specifically on sparse MoE architectures because expert-routing distributions constitute the primary object of analysis. Dense transformers do not expose an explicit routing mechanism, making direct comparisons of routing specialisation impossible. While dense baselines remain useful for behavioural benchmarking, they cannot provide the routing-level signals studied in this work.

While the present study focuses on identifying category-dependent routing behaviour, future work could perform expert masking or routing interventions to determine whether the identified expert preferences play a causal role in linguistic processing.

\bibliography{custom}

\appendix

\section{Probe Token Taxonomy}
\label{app:probe}

\subsection*{English lexical probes}

A token $\tau$ is classified as English lexical if either condition holds:
\begin{enumerate}
  \item \textbf{Irregular past-tense verb.} $\textsc{upos}(\tau)=\texttt{VERB}$, $\texttt{Tense=Past}\in\textsc{feats}(\tau)$, $\texttt{VerbForm=Fin}\in\textsc{feats}(\tau)$, and $\textsc{lemma}(\tau)\in\mathcal{V}^{\text{EN}}_{\text{irr}}$.
  \item \textbf{Irregular plural noun.} $\textsc{upos}(\tau)=\texttt{NOUN}$ and $\textsc{text}(\tau)\in\mathcal{N}^{\text{EN}}_{\text{irr}}$.
\end{enumerate}

\subsection*{German lexical probes}

German lexical probes are operationalized as item-specific or lexically exceptional forms:
\begin{enumerate}
  \item \textbf{Strong past-tense verbs.} $\textsc{upos}(\tau)=\texttt{VERB}$, $\texttt{Tense=Past}\in\textsc{feats}(\tau)$, $\texttt{VerbForm=Fin}\in\textsc{feats}(\tau)$, and $\textsc{lemma}(\tau)\in\mathcal{V}^{\text{DE}}_{\text{irr}}$.
  \item \textbf{Suppletive or highly irregular finite paradigm forms.} $\textsc{text}(\tau)\in\mathcal{S}_{\text{DE}}$ and $\textsc{upos}(\tau)\in\{\texttt{VERB},\texttt{AUX}\}$.
  \item \textbf{Irregular plural nouns.} $\textsc{upos}(\tau)=\texttt{NOUN}$, $\texttt{Number=Plur}\in\textsc{feats}(\tau)$, and $\textsc{text}(\tau)\in\mathcal{N}^{\text{DE}}_{\text{irr}}$.
  \item \textbf{Gender-marked determiners.} $\textsc{upos}(\tau)=\texttt{DET}$ with nominative singular definite or indefinite gender-marked forms. These probes are grouped with lexical probes because they depend on lexically specified noun-gender distinctions, even though they also involve morphosyntactic marking.
\end{enumerate}

\subsection*{English grammatical probes}
\begin{enumerate}
  \item \textbf{Regular past-tense verb.} $\textsc{upos}(\tau)=\texttt{VERB}$, $\texttt{Tense=Past}\in\textsc{feats}(\tau)$, $\texttt{VerbForm=Fin}\in\textsc{feats}(\tau)$, and $\textsc{lemma}(\tau)\notin\mathcal{V}^{\text{EN}}_{\text{irr}}$.
  \item \textbf{Auxiliary agreement.} $\textsc{upos}(\tau)=\texttt{AUX}$ with number-marked auxiliary agreement.
\end{enumerate}

\subsection*{German grammatical probes}
\begin{enumerate}
  \item \textbf{Weak past-tense verb.} $\textsc{upos}(\tau)=\texttt{VERB}$, $\texttt{Tense=Past}\in\textsc{feats}(\tau)$, $\texttt{VerbForm=Fin}\in\textsc{feats}(\tau)$, $\textsc{lemma}(\tau)\notin\mathcal{V}^{\text{DE}}_{\text{irr}}$, and the form is not part of the suppletive set.
  \item \textbf{Auxiliary agreement.} $\textsc{upos}(\tau)=\texttt{AUX}$ with number-marked auxiliary agreement.
\end{enumerate}

\subsection*{Syntactic probes}
\begin{enumerate}
  \item $\textsc{upos}(\tau)=\texttt{SCONJ}$, or
  \item $\textsc{deprel}(\tau)\in\{\texttt{advcl},\texttt{ccomp},\texttt{xcomp},\texttt{acl},\texttt{csubj}\}$.
\end{enumerate}

These labels should be interpreted as operational probe families rather than perfectly discrete theoretical classes. Their purpose is to test whether MoE routing is sensitive to broad linguistically motivated contrasts under a consistent parser-based annotation scheme.

\section{Full Curriculum Schedule}
\label{app:curriculum}

\subsection*{Category sampling}
Within every epoch, lexical, grammatical, and syntactic sentences are sampled uniformly and shuffled randomly.

\subsection*{Forward curriculum}
\begin{table}[h]
\centering\small
\begin{tabular}{lcc}
\toprule
\textbf{Epochs} & \textbf{$\lambda_{\text{DE}}$} & \textbf{Description} \\
\midrule
1--8   & 0.00 & EN only \\
9      & 0.20 & L2 introduction \\
10     & 0.25 & \\
11     & 0.30 & \\
12     & 0.35 & \\
13     & 0.40 & \\
14     & 0.45 & \\
15--16 & 0.50 & Stable bilingual \\
\bottomrule
\end{tabular}
\caption{Forward curriculum L2 exposure schedule.}
\label{tab:l2schedule}
\end{table}

\section{Hyperparameter Details}
\label{app:hparams}

\begin{tabular}{ll}
\toprule
\textbf{Hyperparameter} & \textbf{Value} \\
\midrule
Layers              & 8 \\
Embedding size      & 512 \\
Attention heads     & 8 \\
Feed-forward size   & 2048 \\
MoE layers          & 1, 3, 5, 7 \\
Experts $N$         & 8 \\
Top-$k$             & 3 \\
Load-balance $\alpha$ & 0.01 \\
Vocabulary          & mBERT WordPiece, $\sim$119k \\
Optimizer           & AdamW \\
Peak LR             & $3 \times 10^{-4}$ \\
Warmup steps        & 500 \\
Batch size          & 8 sequences \\
Gradient clip       & 1.0 \\
Epochs              & 16 \\
\bottomrule
\end{tabular}

\subsection*{Computing Infrastructure and Budget}

All training, validation, and routing-analysis experiments were conducted on NVIDIA T4 GPUs and NVIDIA RTX A6000 GPUs. The primary bilingual MoE model used in the study contains 86.4M parameters. Each training condition was run for 16 epochs with 300,000 sentence instances per epoch, corresponding to 4.8M sentence instances per condition. Validation, routing extraction, and downstream statistical analyses were performed from saved checkpoints and held-out routing logs on the same hardware class. Depending on hardware availability, experiments were conducted on either NVIDIA T4 or NVIDIA RTX A6000 GPUs. Training time therefore varied across runs, ranging from approximately 14–28 GPU-hours per condition

\section{Additional Routing Analysis Details}
\label{app:routing}

\subsection*{Permutation testing}
Category labels are shuffled at the sentence level for 1,000 permutations per layer. Empirical $p$-values are computed as the fraction of permutations with $I(R;C)\geq I_{\text{observed}}$. This tests whether the observed routing-category dependence is larger than would be expected under category-independent routing.

\section{OpenAI Generation Details, Validation Parsing, and Prompt Examples}
\label{app:openai_generation}

The held-out validation sentences were generated with the OpenAI Responses API using the Python OpenAI client. In the generation script, the default model was \texttt{GPT-5.1}, configurable through the \texttt{OPENAI\_MODEL} environment variable. Requests were made with \texttt{max\_output\_tokens=420}, \texttt{reasoning.effort="none"}, \texttt{store=false}, request timeout \texttt{40s}, and \texttt{max\_retries=0}. The API call also included a fixed instruction string: \textit{``Follow the output format exactly. Return only the requested lines and nothing else.''}

Generated candidates were then automatically parsed and validated with Stanza using the processors \texttt{tokenize}, \texttt{pos}, \texttt{lemma}, and \texttt{depparse}. For each generated $S^{+}$ sentence, the pipeline checked that the marked probe token occurred exactly once, that the output matched the required subtype inventory, and that the parsed probe token satisfied the subtype-specific constraints used in dataset construction (e.g., finite past-tense verb, plural noun, auxiliary, or subordinating conjunction, depending on the subtype). Additional filters removed malformed outputs, duplicates, and subtype violations before acceptance into the held-out set.

For the final validation pair construction, the ungrammatical sentence $S^{-}$ was obtained deterministically from the validated $S^{+}$ sentence by editing only the probe token or deleting the probe in the syntactic conjunction cases. The transformation rule depended on subtype. For example, English irregular past forms were replaced with regularized forms, English regular past-tense verbs were replaced with their lemma/base form, auxiliary agreement probes were swapped to mismatching number forms, German strong verbs were weak-regularized, German suppletive forms were replaced with present/agreement-incompatible alternatives, and subordinating conjunction probes were removed to create a syntactically degraded variant. This yielded minimal-pair contrasts in which $S^{+}$ remained parser-validated and $S^{-}$ differed by a controlled probe-level manipulation.

Below we provide two illustrative prompt examples adapted directly from the generation templates used in the pipeline.

\paragraph{Prompt Example 1: English irregular verb probe.}
\begin{quote}\small
Write exactly 10 EN sentences.

Subtype: irregular\_verb \\
Preferred lemmas: become, draw, awake, speak, throw, choose, write, drive, sing \\
Avoid recent probes: none

Rules:
- Output exactly 10 lines, no more and no fewer.
- One sentence per line.
- 6--12 words per sentence.
- Prefer a different listed lemma on each line.
- Mark target as [PROBE: word].
- Prefer the correct irregular simple-past form of one listed lemma.
- Probe must be the main verb.
- Use simple past only.
- No auxiliaries or participles with the probe.
- Use the real irregular past form, not the base form.
- Examples: become $\rightarrow$ became, awake $\rightarrow$ awoke, draw $\rightarrow$ drew.
- Prefer the listed lemmas strongly.
- If one listed lemma feels awkward, use another listed lemma.
- No numbering, bullets, quotes, parentheses, or extra text.
- Bare sentence text only.
\end{quote}

\paragraph{Prompt Example 2: German subordinating conjunction probe.}
\begin{quote}\small
Write exactly 10 DE sentences.

Subtype: sconj \\
Preferred subordinating conjunctions: weil, obwohl, bevor, nachdem, falls, sobald, damit, dass \\
Avoid recent probes: none

Rules:
- Output exactly 10 lines, no more and no fewer.
- One sentence per line.
- 6--12 words per sentence.
- Prefer a different listed conjunction on each line.
- Mark target as [PROBE: word].
- Prefer one listed subordinating conjunction exactly as shown.
- Use the probe as a true subordinating conjunction with UPOS=SCONJ.
- Build a clear subordinate-clause construction.
- Keep the sentence fully grammatical as written.
- Avoid coordinators, adverbs, or discourse markers.
- No numbering, bullets, quotes, parentheses, or extra text.
- Bare sentence text only.
\end{quote}

\subsection*{Parsing and Validation Tools}

Parser-based probe annotation and validation were performed with Stanza. We used the Stanza pipelines for English and German with the processors \texttt{tokenize}, \texttt{pos}, \texttt{lemma}, and \texttt{depparse}. These pipeline outputs were used to identify probe tokens, verify subtype constraints, and validate generated held-out examples.

\section{Dataset Diversity Statistics}
\label{app:diversity}

To address concerns regarding dataset health and template collapse, we provide diversity statistics for the diagnostic set. All reported probe sets maintain 100\% sentence-pair uniqueness.

\begin{table}[ht]
\centering
\small
\setlength{\tabcolsep}{4pt}
\begin{tabular}{lccc}
\toprule
Probe Set & Records & Tokens & Uniqueness \\
\midrule
EN Irregular Verb   & 1500 & 46 & 100\% \\
EN Irregular Plural & 1500 & 30 & 100\% \\
DE Strong Verb      & 500  & 35 & 100\% \\
EN SCONJ            & 3000 & 21 & 100\% \\
EN Aux Agreement    & 1500 & 17 & 100\% \\
\bottomrule
\end{tabular}
\caption{Illustrative diagnostic probe diversity statistics.}
\label{tab:probe_diversity}
\end{table}

\section{Ablation Controls}
\label{app:ablations}

The appendix currently includes four additional controls beyond the main forward and no-curriculum MoE comparisons: random routing, frozen routing, top-$k$, and load-balancing ($\alpha=0.005$) ablations under the forward curriculum. These controls are intended to distinguish curriculum effects from generic sparse-routing effects and to test how sensitive the observed specialisation is to router flexibility and routing sparsity.

\begin{table}[ht]
\centering
\small
\setlength{\tabcolsep}{4pt}
\begin{tabular}{lccc}
\toprule
\textbf{Ablation} & \textbf{Combined} & \textbf{EN} & \textbf{DE} \\
\midrule
Random routing & 0.0018 & 0.0029 & 0.0057 \\
Frozen routing & 0.0274 & 0.0647 & 0.0140 \\
Load balancing ($\alpha=0.005$) & 0.1482 & 0.3288 & 0.1042 \\
Forward top-$k=2$ & 0.1100 & 0.1767 & 0.1349 \\
Forward top-$k=4$ & 0.2382 & 0.4107 & 0.1144 \\
\bottomrule
\end{tabular}
\caption{Routing-control and sparsity ablations evaluated on the held-out validation set (MI@L5). Random routing collapses routing-category dependence toward near-null values, indicating that specialisation does not arise from architectural sparsity alone. Freezing router parameters preserves weaker but non-zero specialisation. Lower load balancing increases routing-category dependence, while the routing-sparsity ablations show that top-$k=2$ yields a similar but slightly lower pooled MI than the main model, whereas top-$k=4$ produces substantially stronger category-conditioned routing, especially for English probes.}
\label{tab:ablation_controls}
\end{table}

The random-routing condition replaces learned router assignments with uniformly sampled expert selection while preserving the underlying expert parameters. As expected, routing-category mutual information collapses to near-zero values, indicating that the observed specialisation patterns require adaptive routing rather than arising from architectural sparsity alone.

The frozen-routing condition initializes routing normally but prevents subsequent router optimization during training. Although specialisation is substantially weaker than in the fully trainable model, non-trivial routing-category dependence remains, suggesting that expert differentiation can emerge through changes in token representations interacting with fixed routing boundaries.

Load-balancing strength additionally exerts a strong influence on specialisation structure. Reducing the auxiliary load-balancing coefficient to $\alpha=0.005$ markedly increases routing-category dependence, indicating that weaker balancing constraints permit stronger expert partitioning. However, this increase coincides with increasingly skewed expert utilisation, suggesting a trade-off between specialisation strength and balanced expert participation.

Routing sparsity also affects specialisation, but not in the initially expected direction. Reducing expert selection from top-$k=3$ to top-$k=2$ leaves pooled routing-category dependence at a similar level, with a slight decrease overall. In contrast, increasing routing breadth to top-$k=4$ yields a substantially stronger routing-category association, especially for English probes. In the present runs, broader routing therefore coincides with stronger category-conditioned separation rather than weaker specialisation.

\subsection{Seed Replication Detail}
\label{app:seedreplication}

Table~\ref{tab:seed_replication} gives the full per-seed, per-language breakdown underlying the seed-replication discussion in Section~5.6. The pooled structure (no-curriculum $>$ forward at layer 5) and the curriculum's smaller cross-language asymmetry reproduce across both seeds; the identity of the language that dominates under the no-curriculum condition does not.

\begin{table}[h]
\centering
\small
\setlength{\tabcolsep}{4pt}
\begin{tabular}{lcccc}
\toprule
 & \multicolumn{2}{c}{Forward} & \multicolumn{2}{c}{No-Curr} \\
MI@L5 & seed 42 & seed 50 & seed 42 & seed 50 \\
\midrule
Combined & 0.115 & 0.089 & 0.260 & 0.113 \\
English  & 0.103 & 0.148 & 0.646 & 0.078 \\
German   & 0.144 & 0.063 & 0.112 & 0.213 \\
$|$EN$-$DE$|$ & 0.042 & 0.084 & 0.534 & 0.135 \\
\bottomrule
\end{tabular}
\caption{Layer-5 routing-category MI at both training seeds.}
\label{tab:seed_replication}
\end{table}

\subsection{Frequency-Binned Lexical Analysis}

To examine whether lexical routing specialisation is driven primarily by token frequency, English lexical probes were grouped into frequency bins using curriculum-weighted lemma frequencies. Mutual information was then recomputed using only lexical irregular-versus-regular contrasts within each frequency bin.

\begin{table}[h]
\centering
\small
\begin{tabular}{lcc}
\toprule
Frequency Bin & Forward MI@L5 & No-Curr MI@L5 \\
\midrule
Frequent & 0.0149 & 0.0055 \\
Medium   & 0.0206 & 0.0976 \\
Rare     & 0.2261 & 0.2494 \\
\bottomrule
\end{tabular}
\caption{English lexical irregular-versus-regular routing mutual information at layer 5 after binning probe lemmas by curriculum-weighted frequency.}
\label{tab:freq_bin_lexical}
\end{table}

Mutual information remains non-zero across all frequency bins in both training conditions. Notably, the strongest routing-category association is observed for rare lexical items rather than the most frequent items. This suggests that the lexical routing effect cannot be explained solely by the highest-frequency lexical forms.

\begin{table}[h]
\centering
\small
\begin{tabular}{lcc}
\toprule
Layer & Forward $I(R;C \mid L)$ & No-Curr $I(R;C \mid L)$ \\
\midrule
1 & 0.0935 & 0.2584 \\
3 & 0.0639 & 0.1084 \\
5 & 0.1235 & 0.3787 \\
7 & 0.0312 & 0.0895 \\
\bottomrule
\end{tabular}
\caption{Conditional mutual information between routing and linguistic category given language, $I(R;C \mid L)$, across routed layers. In both the forward-curriculum and no-curriculum conditions, conditional MI remains clearly non-zero and permutation-significant at all routed layers ($p=0.000999$), indicating that category-sensitive routing is not reducible to language separation alone.}
\label{tab:conditional_mi}
\end{table}

We additionally examined the evolution of routing specialisation during training using epoch-wise routing logs. These analyses are provided here as supplementary evidence.

\subsection{Routing Specialisation Dynamics During Training}

\begin{figure}[ht]
    \centering
    \includegraphics[width=\linewidth]{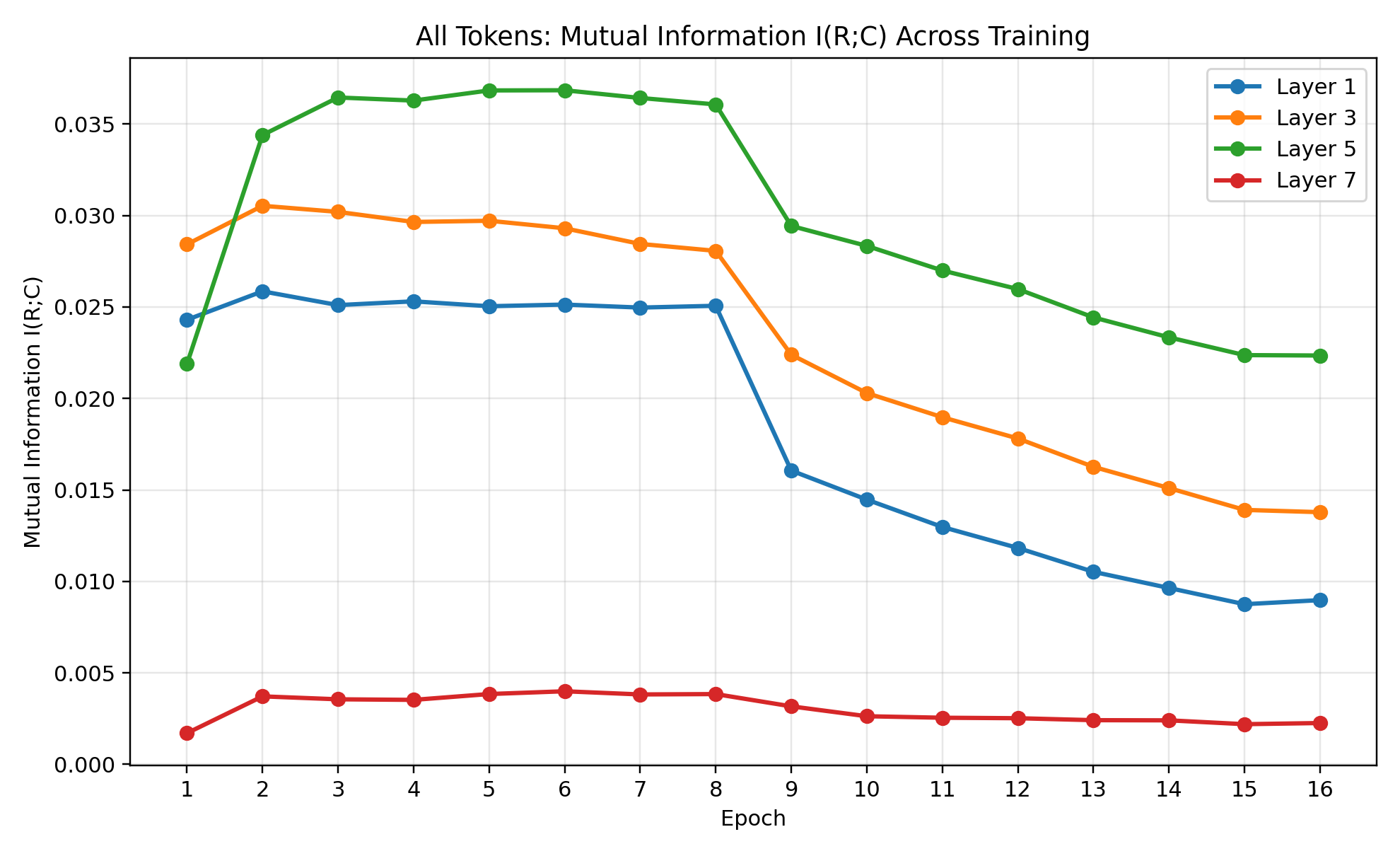}
    \caption{Evolution of routing-category mutual information across training epochs for the routed layers.}
    \label{fig:training_mi}
\end{figure}

To analyse routing specialisation throughout training, mutual information and entropy were computed from the routing logs at checkpoints from each training epoch. Figure~\ref{fig:training_mi} shows that category-dependent routing emerges rapidly during the early stages of training and stabilises over the following epochs.

The relative ordering of the routed layers remains largely consistent throughout training, with layer 5 exhibiting the highest mutual information across all epochs. The figure also shows a decline in category-dependent routing after epoch 8. Following this transition, the mutual information decreases across all routed layers before stabilising again during the later stages of training.

\begin{figure}[ht]
    \centering
    \includegraphics[width=\linewidth]{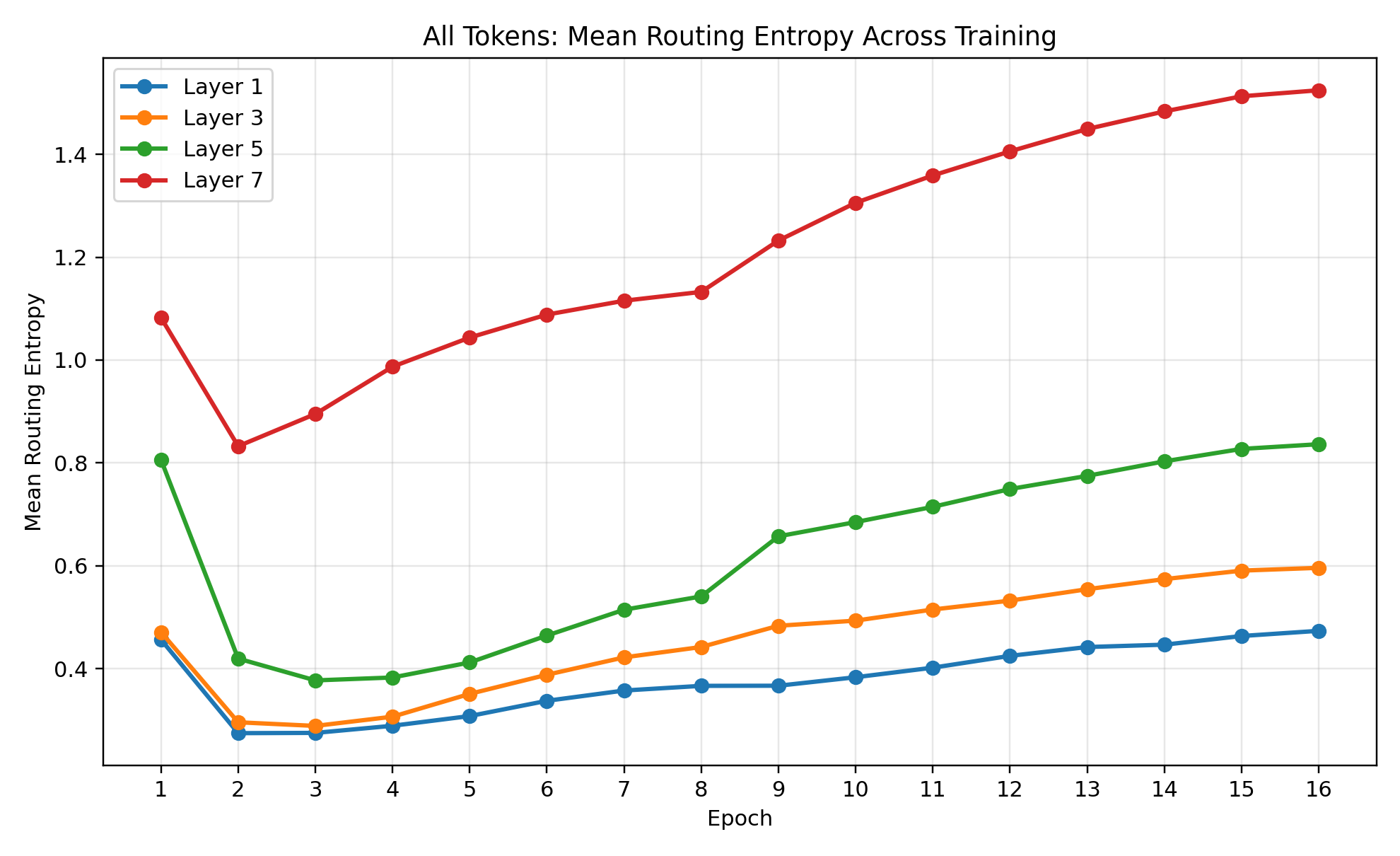}
    \caption{Evolution of routing entropy across training epochs for the routed layers.}
    \label{fig:training_entropy}
\end{figure}

Figure~\ref{fig:training_entropy} presents the routing entropy across the routed layers throughout training. A complementary trend is observed between mutual information and entropy. As mutual information decreases after epoch 8, routing entropy increases across all routed layers during the same period. This indicates that expert utilisation becomes progressively more distributed while category-dependent routing behaviour remains present.

\end{document}